# Local Consistency and SAT-Solvers

**Peter Jeavons**　　　　　　　　　　　　　　　　　　　Peter.Jeavons@cs.ox.ac.uk
**Justyna Petke**　　　　　　　　　　　　　　　　　　　Justyna.Petke@cs.ox.ac.uk
*Department of Computer Science, University of Oxford*
*Wolfson Building, Parks Road, Oxford, OX1 3QD, UK*

## Abstract

Local consistency techniques such as $k$-consistency are a key component of specialised solvers for constraint satisfaction problems. In this paper we show that the power of using $k$-consistency techniques on a constraint satisfaction problem is precisely captured by using a particular inference rule, which we call negative-hyper-resolution, on the standard direct encoding of the problem into Boolean clauses. We also show that current clause-learning SAT-solvers will discover in expected polynomial time any inconsistency that can be deduced from a given set of clauses using negative-hyper-resolvents of a fixed size. We combine these two results to show that, without being explicitly designed to do so, current clause-learning SAT-solvers efficiently simulate $k$-consistency techniques, for all fixed values of $k$. We then give some experimental results to show that this feature allows clause-learning SAT-solvers to efficiently solve certain families of constraint problems which are challenging for conventional constraint-programming solvers.

## 1. Introduction

One of the oldest and most central ideas in constraint programming, going right back to Montanari's original paper in 1974, is the idea of using *local consistency* techniques to prune the search space (Bessière, 2006). The idea of arc-consistency was introduced by Mackworth (1977), and generalised to $k$-consistency by Freuder (1978). Modern constraint solvers generally employ specialised propagators to prune the domains of variables to achieve some form of generalised arc-consistency, but typically do *not* attempt to enforce higher levels of consistency, such as path-consistency.

By contrast, the software tools developed to solve propositional satisfiability problems, known as SAT-solvers, generally use logical inference techniques, such as unit propagation and clause-learning, to prune the search space.

One of the most surprising empirical findings of the last few years has been the remarkably good performance of general SAT-solvers in solving constraint satisfaction problems. To apply such tools to a constraint satisfaction problem one first has to translate the instance into a set of clauses using some form of Boolean encoding (Tamura, Taga, Kitagawa, & Banbara, 2009; Walsh, 2000). Such encoding techniques tend to obscure the structure of the original problem, and may introduce a very large number of Boolean variables and clauses to encode quite easily-stated constraints. Nevertheless, in quite a few cases, such approaches have out-performed more traditional constraint-solving tools (van Dongen, Lecoutre, & Roussel, 2008, 2009; Petke & Jeavons, 2009).

In this paper we draw on a number of recent analytical approaches to try to account for the good performance of general SAT-solvers on many forms of constraint problems. Building on the results of Atserias, Bulatov, and Dalmau (2007), Atserias and Dalmau (2008), and Hwang and Mitchell (2005), we show that the power of using $k$-consistency techniques in a constraint problem is precisely captured by using a single inference rule in a standard Boolean encoding of that problem. We refer to this inference rule as *negative-hyper-resolution*, and show that any conclusions deduced by enforcing $k$-consistency can be deduced by a sequence of negative-hyper-resolution inferences involving Boolean clauses in the original instance and negative-hyper-resolvents with at most $k$ literals. Furthermore, by using the approach of Atserias, Fichte, and Thurley (2011), and Pipatsrisawat and Darwiche (2009), we show that current clause-learning SAT-solvers will mimic the effect of such deductions in polynomial expected time, even with a random branching strategy. Hence we show that, although they are not explicitly designed to do so, running a clause-learning SAT-solver on a straightforward encoding of a constraint problem efficiently simulates the effects of enforcing $k$-consistency for *all* values of $k$.

## 2. Preliminaries

In this section we give some background and definitions that will be used throughout the rest of the paper.

### 2.1 Constraint Satisfaction Problems and $k$-Consistency

**Definition 1** *An instance of the* **Constraint Satisfaction Problem** *(CSP) is specified by a triple* $(V, D, C)$*, where*

- $V$ *is a finite set of* variables*;*

- $D = \{D_v \mid v \in V\}$ *where each set* $D_v$ *is the set of possible values for the variable* $v$*, called the* domain *of* $v$*;*

- $C$ *is a finite set of* constraints*. Each constraint in* $C$ *is a pair* $(R_i, S_i)$ *where*

  - $S_i$ *is an ordered list of* $m_i$ *variables, called the constraint* scope*;*

  - $R_i$ *is a relation over* $D$ *of arity* $m_i$*, called the constraint* relation*.*

Given any CSP instance $(V, D, C)$, a *partial assignment* is a mapping $f$ from some subset $W$ of $V$ to $\bigcup D_v$ such that $f(v) \in D_v$ for all $v \in W$. A partial assignment *satisfies the constraints* of the instance if, for all $(R, (v_1, v_2, \ldots, v_m)) \in C$ such that $v_j \in W$ for $j = 1, 2, \ldots, m$, we have $(f(v_1), f(v_2) \ldots, f(v_m)) \in R$. A partial assignment that satisfies the constraints of an instance is called a *partial solution*[1] to that instance. The set of variables on which a partial assignment $f$ is defined is called the domain of $f$, and denoted $Dom(f)$. A partial solution $g$ *extends* a partial solution $f$ if $Dom(g) \supseteq Dom(f)$ and $g(v) = f(v)$ for all $v \in Dom(f)$. A partial solution with domain $V$ is called a *solution*. One way to derive new information about a CSP instance, which may help to determine whether or not it has a solution, is to use some form of constraint propagation to enforce some level of *local consistency* (Bessière, 2006). For example, it is possible to use the notion of *k-consistency*, defined below. We note that there are several different but equivalent ways to define and enforce $k$-consistency described in the literature (Bessière, 2006; Cooper, 1989; Freuder, 1978). Our presentation follows that of Atserias et al. (2007), which is inspired by the notion of existential $k$-pebble games introduced by Kolaitis and Vardi (2000).

---

1. Note that not all partial solutions extend to solutions.

**Definition 2** *(Atserias et al., 2007) For any CSP instance $P$, the $k$-**consistency closure** of $P$ is the set $H$ of partial assignments which is obtained by the following algorithm:*

1. *Let $H$ be the collection of all partial solutions $f$ of $P$ with $|Dom(f)| \leq k+1$;*

2. *For every $f \in H$ with $|Dom(f)| \leq k$ and every variable $v$ of $P$, if there is no $g \in H$ such that $g$ extends $f$ and $v \in Dom(g)$, then remove $f$ and all its extensions from $H$;*

3. *Repeat step 2 until $H$ is unchanged.*

Note that computing the $k$-consistency closure according to this definition corresponds precisely to enforcing *strong $(k+1)$-consistency* according to the definitions given by Bessière (2006), Cooper (1989), and Freuder (1978).

Throughout this paper, we shall assume that the domain of possible values for each variable in a CSP instance is finite. It is straightforward to show that for any fixed $k$, and any fixed maximum domain size, the $k$-consistency closure of an instance $P$ can be computed in polynomial time (Atserias et al., 2007; Cooper, 1989).

Note that any solution to $P$ must extend some element of the $k$-consistency closure of $P$. Hence, if the $k$-consistency closure of $P$ is empty, for some $k$, then $P$ has no solutions. The converse is not true in general, but it holds for certain special cases, such as the class of instances whose structure has tree-width bounded by $k$ (Atserias et al., 2007), or the class of instances whose constraint relations are "0/1/all" relations, as defined in Cooper, Cohen, and Jeavons (1994), or "connected row-convex" relations, as defined in Deville, Barette, and Hentenryck (1997). For these special kinds of instances it is possible to determine in polynomial time whether or not a solution exists simply by computing the $k$-consistency closure, for an appropriate choice of $k$. Moreover, if a solution exists, then it can be constructed in polynomial time by selecting each variable in turn, assigning each possible value, re-computing the $k$-consistency closure, and retaining an assignment that gives a non-empty result.

The following result gives a useful condition for determining whether the $k$-consistency closure of a CSP instance is empty.

**Lemma 1** *(Kolaitis & Vardi, 2000) The $k$-consistency closure of a CSP instance $P$ is non-empty if and only if there exists a non-empty family $H$ of partial solutions to $P$ such that:*

1. *If $f \in H$, then $|Dom(f)| \leq k+1$;*

2. *If $f \in H$ and $f$ extends $g$, then $g \in H$;*

3. *If $f \in H$, $|Dom(f)| \leq k$, and $v \notin Dom(f)$ is a variable of $P$, then there is some $g \in H$ such that $g$ extends $f$ and $v \in Dom(g)$.*

A set of partial solutions $H$ satisfying the conditions described in Lemma 1 is sometimes called a *strategy* for the instance $P$ (Barto & Kozik, 2009; Kolaitis & Vardi, 2000).

## 2.2 Encoding a CSP Instance as a Propositional Formula

One possible approach to solving a CSP instance is to encode it as a propositional formula over a suitable set of Boolean variables, and then use a program to decide the satisfiability of that formula. Many such programs, known as SAT-solvers, are now available and can often efficiently handle problems with thousands, or sometimes even millions, of Boolean variables (Zhang & Malik, 2002).

Several different ways of encoding a CSP instance as a propositional formula have been proposed (Prestwich, 2009; Tamura et al., 2009; Walsh, 2000).

Here we consider one common family of encodings, known as *sparse encodings* (this term was introduced in Hoos, 1999). For any CSP instance $P = (V, D, C)$, a sparse encoding introduces a set of Boolean variables of the form $x_{vi}$ for each $v \in V$ and each $i \in D_v$. The Boolean variable $x_{vi}$ is assigned *True* if and only if the original variable $v$ is assigned the value $i$. We will say that a partial assignment $f$ *falsifies* a clause $C$ if $C$ consists entirely of literals of the form $\neg x_{vf(v)}$, for variables $v \in Dom(f)$. Otherwise, we will say that a partial assignment $f$ *satisfies* a clause $C$.

**Example 1** *Let $P$ be a CSP instance such that $V = \{u, v, w\}$, $D_u = D_v = \{0, 1\}, D_w = \{0, 1, 2\}$ and $C$ contains a single ternary constraint with scope $(u, v, w)$ specifying that $u \leq v < w$. A sparse encoding of $P$ will introduce seven Boolean variables:*

$$x_{u0}, \ x_{u1}, \ x_{v0}, \ x_{v1}, \ x_{w0}, \ x_{w1}, \ x_{w2}.$$

Sparse encodings usually contain certain clauses known as *at-least-one* and *at-most-one* clauses, to ensure that each variable $v$ is assigned a value, say $i$, and that no other value, $j \neq i$, is assigned to $v$. The at-least-one clauses are of the form $\bigvee_{i \in D_v} x_{vi}$ for each variable $v$. The at-most-one clauses can be represented as a set of binary clauses $\neg x_{vi} \vee \neg x_{vj}$ for all $i, j \in D_v$ with $i \neq j$.

**Example 2** *In the case of the CSP instance from Example 1 the at-least-one clauses are:*

$$x_{u0} \vee x_{u1}, \ x_{v0} \vee x_{v1}, \ x_{w0} \vee x_{w1} \vee x_{w2}$$

*The at-most-one clauses are:*

$$\neg x_{u0} \vee \neg x_{u1}, \ \neg x_{v0} \vee \neg x_{v1}, \ \neg x_{w0} \vee \neg x_{w1}, \ \neg x_{w0} \vee \neg x_{w2}, \ \neg x_{w1} \vee \neg x_{w2}$$

The various different sparse encodings differ in the way they encode the constraints of a CSP instance. Two methods are most commonly used. The first one encodes the *disallowed* variable assignments - the so-called *conflicts* or *no-goods*. The **direct encoding** (Prestwich, 2009), for instance, generates a clause $\bigvee_{v \in S} \neg x_{vf(v)}$ for each partial assignment $f$ that does *not* satisfy the constraint $(R, S) \in C$. Using the direct encoding, the ternary constraint from Example 1 would be encoded by the following clauses:

$$\neg x_{u0} \vee \neg x_{v0} \vee \neg x_{w0},$$
$$\neg x_{u0} \vee \neg x_{v1} \vee \neg x_{w0},$$
$$\neg x_{u0} \vee \neg x_{v1} \vee \neg x_{w1},$$
$$\neg x_{u1} \vee \neg x_{v0} \vee \neg x_{w0},$$

$$\neg x_{u1} \vee \neg x_{v0} \vee \neg x_{w1},$$
$$\neg x_{u1} \vee \neg x_{v0} \vee \neg x_{w2},$$
$$\neg x_{u1} \vee \neg x_{v1} \vee \neg x_{w0},$$
$$\neg x_{u1} \vee \neg x_{v1} \vee \neg x_{w1}.$$

Another way of translating constraints into clauses is to encode the *allowed* variable assignments - the so-called *supports*. This has been used as the basis for an encoding of binary CSP instances, known as the **support encoding** (Gent, 2002), defined as follows. For each pair of variables $v, w$ in the scope of some constraint, and each value $i \in D_v$, the support encoding will contain the clause $\neg x_{vi} \vee \bigvee_{j \in A} x_{wj}$, where $A \subseteq D_w$ is the set of values for the variable $w$ which are compatible with the assignment $v = i$, according to the constraint.

Note that the support encoding is defined for binary CSP instances only. However, some non-binary constraints can be decomposed into binary ones without introducing any new variables. For instance, the ternary constraint from Example 1 can be decomposed into two binary constraints specifying that $u \leq v$ and $v < w$. Using the support encoding, these binary constraints would then be represented by the following clauses:

$$\neg x_{u0} \vee x_{v0} \vee x_{v1}, \ \neg x_{u1} \vee x_{v1}, \ \neg x_{v0} \vee x_{u0}, \ \neg x_{v1} \vee x_{u0} \vee x_{u1},$$
$$\neg x_{v0} \vee x_{w1} \vee x_{w2}, \ \neg x_{v1} \vee x_{w2}, \ \neg x_{w0}, \ \neg x_{w1} \vee x_{v0}, \ \neg x_{w2} \vee x_{v0} \vee x_{v1}.$$

### 2.3 Inference Rules

Given any set of clauses we can often deduce further clauses by applying certain *inference rules*. For example, if we have two clauses of the form $C_1 \vee x$ and $C_2 \vee \neg x$, for some (possibly empty) clauses $C_1, C_2$, and some variable $x$, then we can deduce the clause $C_1 \vee C_2$. This form of inference is known as *propositional resolution*; the resultant clause is called the *resolvent* (Robinson, 1965).

In the next section, we shall establish a close connection between the $k$-consistency algorithm and a form of inference called negative-hyper-resolution (Büning & Lettmann, 1999), which we define as follows:

**Definition 3** *If we have a collection of clauses of the form $C_i \vee \neg x_i$, for $i = 1, 2, \ldots, r$, and a clause $C_0 \vee x_1 \vee x_2 \vee \cdots \vee x_r$, where each $x_i$ is a Boolean variable, and $C_0$ and each $C_i$ is a (possibly empty) disjunction of negative literals, then we can deduce the clause $C_0 \vee C_1 \vee \cdots \vee C_r$.*

*We call this form of inference **negative-hyper-resolution** and the resultant clause $C_0 \vee C_1 \vee \cdots \vee C_r$ the negative-hyper-resolvent.*

In the case where $C_0$ is empty, the negative-hyper-resolution rule is equivalent to the nogood resolution rule described by Hwang and Mitchell (2005) as well as the H5-$k$ rule introduced by de Kleer (1989) and the nogood recording scheme described by Schiex and Verfaillie (1993).

Note that the inference obtained by negative-hyper-resolution can also be obtained by a sequence of standard resolution steps. However, the reason for introducing negative-hyper-resolution is that it allows us to deduce the clauses we need in a single step without needing to introduce intermediate clauses (which may contain up to $r - 1$ more literals than the negative-hyper-resolvent). By restricting the size of the clauses we use in this way we are able to obtain better performance bounds for SAT-solvers in the results below.

**Example 3** *Assume we have a collection of clauses of the form $C_i \vee \neg x_i$, for $i = 1, 2, \ldots, r$, and a clause $C_0 \vee x_1 \vee x_2 \vee \cdots \vee x_r$, as specified in Definition 3, where each $C_i = C_0$. The negative-hyper-resolvent of this set of clauses is $C_0$.*

*The clause $C_0$ can also be obtained by a sequence of standard resolution steps, as follows. First resolve $C_0 \vee x_1 \vee x_2 \vee \cdots \vee x_r$ with $C_0 \vee \neg x_r$ to obtain $C_0 \vee x_1 \vee x_2 \vee \cdots \vee x_{r-1}$. Then resolve this with the next clause, $C_0 \vee \neg x_{r-1}$, and so on for the other clauses, until finally we obtain $C_0$. However, in this case the intermediate clause $C_0 \vee x_1 \vee x_2 \vee \cdots \vee x_{r-1}$ contains $r - 1$ more literals than the negative-hyper-resolvent.*

**Example 4** *Note that the no-good clauses in the direct encoding of a binary CSP instance can each be obtained by a single negative-hyper-resolution step from an appropriate support clause in the support encoding together with an appropriate collection of at-most-one clauses. Let $A \subseteq D_w$ be the set of values for the variable $w$ which are compatible with the assignment $v = i$, then the support encoding will contain the clause $C = \neg x_{vi} \vee \bigvee_{j \in A} x_{wj}$. If there are any values $k \in D_w$ which are incompatible with the assignment $v = i$, then we can form the negative-hyper-resolvent of $C$ with the at-most-one clauses $\neg x_{wk} \vee \neg x_{wj}$ for each $j \in A$, to obtain the corresponding no-good clause, $\neg x_{vi} \vee \neg x_{wk}$.*

A negative-hyper-resolution *derivation* of a clause $C$ from a set of initial clauses $\Phi$ is a sequence of clauses $C_1, C_2, \ldots, C_m$, where $C_m = C$ and each $C_i$ follows by the negative-hyper-resolution rule from some collection of clauses, each of which is either contained in $\Phi$ or else occurs earlier in the sequence. The *width* of this derivation is defined to be the maximum size of any of the clauses $C_i$. If $C_m$ is the empty clause, then we say that the derivation is a *negative-hyper-resolution refutation* of $\Phi$.

## 3. $k$-Consistency and Negative-Hyper-Resolution

It has been pointed out by many authors that enforcing local consistency is a form of inference on relations analogous to the use of the resolution rule on clauses (Bacchus, 2007; Bessière, 2006; Hwang & Mitchell, 2005; Rish & Dechter, 2000). The precise strength of the standard resolution inference rule on the direct encoding of a CSP instance was considered in the work of Walsh (2000), where it was shown that *unit* resolution (where one of the clauses being resolved consists of a single literal), corresponds to enforcing a weak form of local consistency known as *forward checking*. Hwang and Mitchell (2005) pointed out that the standard resolution rule with no restriction on clause length is able to simulate all the inferences made by a $k$-consistency algorithm. Atserias and Dalmau (2008) showed that the standard resolution rule restricted to clauses with at most $k$ literals, known as the $k$-resolution rule, can be characterised in terms of the Boolean existential $(k+1)$-pebble game. It follows that on CSP instances with Boolean domains this form of inference corresponds to enforcing $k$-consistency. An alternative proof that $k$-resolution achieves $k$-consistency for instances with Boolean domains is given in the book by Hooker (2006, Thm. 3.22).

Here we extend these results a little, to show that for CSP instances with arbitrary finite domains, applying the negative-hyper-resolution rule on the direct encoding to obtain clauses with at most $k$ literals corresponds precisely to enforcing $k$-consistency. A similar relationship was stated in the work of de Kleer (1989), but a complete proof was not given.

Note that the bound, $k$, that we impose on the size of the negative-hyper-resolvents, is independent of the domain size. In other words, using this inference rule we only need to consider inferred clauses of size at most $k$, even though we make use of clauses in the encoding whose size is equal to the domain size, which may be arbitrarily large.

**Theorem 1** *The $k$-consistency closure of a CSP instance $P$ is empty if and only if its direct encoding as a set of clauses has a negative-hyper-resolution refutation of width at most $k$.*

The proof is broken down into two lemmas inspired by Lemmas 2 and 3 in the work of Atserias and Dalmau (2008).

**Lemma 2** *Let $P$ be a CSP instance, and let $\Phi$ be its direct encoding as a set of clauses. If $\Phi$ has no negative-hyper-resolution refutation of width $k$ or less, then the $k$-consistency closure of $P$ is non-empty.*

*Proof.* Let $V$ be the set of variables of $P$, where each $v \in V$ has domain $D_v$, and let $X = \{x_{vi} \mid v \in V, i \in D_v\}$ be the corresponding set of Boolean variables in $\Phi$. Let $\Gamma$ be the set of all clauses having a negative-hyper-resolution derivation from $\Phi$ of width at most $k$. By the definition of negative-hyper-resolution, every non-empty clause in $\Gamma$ consists entirely of negative literals.

Now let $H$ be the set of all partial assignments for $P$ with domain size at most $k + 1$ that do not falsify any clause in $\Phi \cup \Gamma$ under the direct encoding.

Consider any element $f \in H$. By the definition of $H$, $f$ does not falsify any clause of $\Phi$, so by the definition of the direct encoding, every element of $H$ is a partial solution to $P$. Furthermore, if $f$ extends $g$, then $g$ is also an element of $H$, because $g$ makes fewer assignments than $f$ and hence cannot falsify any additional clauses to $f$.

If $\Phi$ has no negative-hyper-resolution refutation of width at most $k$, then $\Gamma$ does not contain the empty clause, so $H$ contains (at least) the partial solution with empty domain, and hence $H$ is not empty.

Now let $f$ be any element of $H$ with $|Dom(f)| \leq k$ and let $v$ be any variable of $P$ that is not in $Dom(f)$. For any partial assignment $g$ that extends $f$ and has $Dom(g) = Dom(f) \cup \{v\}$ we have that either $g \in H$ or else there exists a clause in $\Phi \cup \Gamma$ that is falsified by $g$. Since $g$ is a partial assignment, any clause $C$ in $\Phi \cup \Gamma$ that is falsified by $g$, must consist entirely of negative literals. Hence the literals of $C$ must either be of the form $\neg x_{wf(w)}$ for some $w \in Dom(f)$, or else $\neg x_{vg(v)}$. Moreover, any such clause must contain the literal $\neg x_{vg(v)}$, or else it would already be falsified by $f$.

Assume, for contradiction, that $H$ does not contain any assignment $g$ that extends $f$ and has $Dom(g) = Dom(f) \cup \{v\}$. In that case, we have that, for each $i \in D_v$, $\Phi \cup \Gamma$ contains a clause $C_i$ consisting of negative literals of the form $\neg x_{wf(w)}$ for some $w \in Dom(f)$, together with the literal $\neg x_{vi}$. Now consider the clause, $C$, which is the negative-hyper-resolvent of these clauses $C_i$ and the at-least-one clause $\bigvee_{i \in D_v} x_{vi}$. The clause $C$ consists entirely of negative literals of the form $\neg x_{wf(w)}$ for some $w \in Dom(f)$, so it has width at most $|Dom(f)| \leq k$, and hence is an element of $\Gamma$. However $C$ is falsified by $f$, which contradicts the choice of $f$. Hence we have shown that for all $f \in H$ with $|Dom(f)| \leq k$, and for all variables $v$ such that $v \notin Dom(f)$, there is some $g \in H$ such that $g$ extends $f$ and $v \in Dom(g)$.

We have shown that $H$ satisfies all the conditions required by Lemma 1, so we conclude that the $k$-consistency closure of $P$ is non-empty. □

**Lemma 3** *Let $P$ be a CSP instance, and let $\Phi$ be its direct encoding as a set of clauses. If the $k$-consistency closure of $P$ is non-empty, then $\Phi$ has no negative-hyper-resolution refutation of width $k$ or less.*

*Proof.* Let $V$ be the set of variables of $P$, where each $v \in V$ has domain $D_v$, and let $X = \{x_{vi} \mid v \in V, i \in D_v\}$ be the corresponding set of Boolean variables in $\Phi$.

By Lemma 1, if the $k$-consistency closure of $P$ is non-empty, then there exists a non-empty set $H$ of partial solutions to $P$ which satisfies the three properties described in Lemma 1.

Now consider any negative-hyper-resolution derivation $\Gamma$ from $\Phi$ of width at most $k$. We show by induction on the length of this derivation that the elements of $H$ do not falsify any clause in the derivation. First we note that the elements of $H$ are partial solutions, so they satisfy all the constraints of $P$, and hence do not falsify any clause of $\Phi$. This establishes the base case. Assume, for induction, that all clauses in the derivation earlier than some clause $C$ are not falsified by any element of $H$.

Note that, apart from the at-least-one clauses, all clauses in $\Phi$ and $\Gamma$ consist entirely of negative literals. Hence we may assume, without loss of generality, that $C$ is the negative-hyper-resolvent of a set of clauses $\Delta = \{C_i \vee \neg x_{vi} \mid i \in D_v\}$ and the at-least-one clause $\bigvee_{i \in D_v} x_{vi}$, for some fixed variable $v$.

If $f \in H$ falsifies $C$, then the literals of $C$ must all be of the form $\neg x_{wf(w)}$, for some $w \in Dom(f)$. Since the width of the derivation is at most $k$, $C$ contains at most $k$ literals, and hence we may assume that $|Dom(f)| \leq k$. But then, by the choice of $H$, there must exist some extension $g$ of $f$ in $H$ such that $v \in Dom(g)$. Any such $g$ will falsify some clause in $\Delta$, which contradicts our inductive hypothesis. Hence no $f \in H$ falsifies $C$, and, in particular, $C$ cannot be empty.

It follows that no negative-hyper-resolution derivation of width at most $k$ can contain the empty clause. □

Note that the proof of Theorem 1 applies to any sparse encoding that contains the at-least-one clauses for each variable, and where all other clauses are purely negative. We will call such an encoding a *negative sparse encoding*. As well as the direct encoding, other negative sparse encodings exist. For example, we may use negative clauses that involve only a subset of the variables in the scope of some constraints (to forbid tuples where all possible extensions to the complete scope are disallowed by the constraint). Another example of a negative sparse encoding is a well-known variant of the direct encoding in which the at-most-one clauses are omitted.

**Corollary 1** *The $k$-consistency closure of a CSP instance $P$ is empty if and only if any negative sparse encoding of $P$ has a negative-hyper-resolution refutation of width at most $k$.*

## 4. Negative-Hyper-Resolution and SAT-Solvers

In this section we adapt the machinery from Atserias et al. (2011), and Pipatsrisawat and Darwiche (2009) to show that for any fixed $k$, the existence of a negative-hyper-resolution refutation of width $k$ is likely to be discovered by a SAT-solver in polynomial-time using standard clause learning and restart techniques, even with a totally random branching strategy.

Note that previous results about the power of clause-learning SAT-solvers have generally assumed an optimal branching strategy (Beame, Kautz, & Sabharwal, 2004; Pipatsrisawat & Darwiche, 2009) - they have shown what solvers are potentially capable of doing, rather than what they are likely to achieve in practice. An important exception is the paper by Atserias et al. (2011), which gives an analysis of likely behaviour, but relies on the existence of a standard resolution proof of bounded width. Here we show that the results of Atserias et al. can be extended to hyper-resolution proofs, which can be shorter and narrower than their associated standard resolution proofs.

We will make use of the following terminology from Atserias et al. (2011). For a clause $C$, a Boolean variable $x$, and a truth value $a \in \{0, 1\}$, the restriction of $C$ by the assignment $x = a$, denoted $C|_{x=a}$, is defined to be the constant $\mathbf{1}$, if the assignment satisfies the clause, or else the clause obtained by deleting from $C$ any literals involving the variable $x$. For any sequence of assignments $S$ of the form $(x_1 = a_1, x_2 = a_2, \ldots, x_r = a_r)$ we write $C|_S$ to denote the result of computing the restriction of $C$ by each assignment in turn. If $C|_S$ is empty, then we say that the assignments in $S$ *falsify* the clause $C$. For a set of clauses $\Delta$, we write $\Delta|_S$ to denote the set $\{C|_S \mid C \in \Delta\} \setminus \{\mathbf{1}\}$.

Most current SAT-solvers operate in the following way (Atserias et al., 2011; Pipatsrisawat & Darwiche, 2009). They maintain a database of clauses $\Delta$ and a current state $S$, which is a partial assignment of truth values to the Boolean variables in the clauses of $\Delta$. A high-level description of the algorithms used to update the clause database and the state, derived from the description given in Atserias et al., is shown in Algorithm 1 (a similar framework, using slightly different terminology, is given in Pipatsrisawat & Darwiche, 2009).

Now consider a run of the algorithm shown in Algorithm 1, started with the initial database $\Delta$, and the empty state $S_0$, until it either halts or discovers a *conflict* (i.e., $\emptyset \in \Delta|_S$). Such a run is called a *complete round started with* $\Delta$, and we represent it by the sequence of states $S_0, \ldots, S_m$, that the algorithm maintains. Note that each state $S_i$ extends the state $S_{i-1}$ by a single assignment to a Boolean variable, which may be either a *decision assignment* or an *implied assignment*.

More generally, a *round* is an initial segment $S_0, S_1, \ldots, S_r$ of a complete round started with $\Delta$, up to a state $S_r$ such that either $\Delta|_{S_r}$ contains the empty clause, or $\Delta|_{S_r}$ does not contain any unit clause. For any clause $C$, we say that a round $S_0, S_1, \ldots, S_r$ satisfies $C$ if $C|_{S_r} = \mathbf{1}$, and we say that the round falsifies $C$ if $C|_{S_r}$ is empty.

If $S_0, S_1, \ldots, S_r$ is a round started with $\Delta$, and $\Delta|_{S_r}$ contains the empty clause, then the algorithm either reports unsatisfiability or learns a new clause: such a round is called *conclusive*. If a round is not conclusive we call it *inconclusive* [2]. Note that if $S_0, S_1, \ldots, S_r$ is an inconclusive round started with $\Delta$, then $\Delta|_{S_r}$ does not contain the empty clause, and does not contain any unit clauses. Hence, for any clause $C \in \Delta$, if $S_r$ falsifies all the literals of $C$ except one, then it must satisfy the remaining literal, and hence satisfy $C$. This property of clauses is captured by the following definition.

---

[2]. Note that a complete round that assigns all variables and reports satisfiability is called inconclusive.

**Definition 4** *(Atserias et al., 2011) Let $\Delta$ be a set of clauses, $C$ a non-empty clause, and $l$ a literal of $C$. We say that $\Delta$ absorbs $C$ at $l$ if every inconclusive round started with $\Delta$ that falsifies $C \setminus \{l\}$ satisfies $C$.*

*If $\Delta$ absorbs $C$ at each literal $l$ in $C$, then we simply say that $\Delta$ **absorbs** $C$.*

Note that a closely related notion is introduced by Pipatsrisawat and Darwiche (2009) for clauses that are *not* absorbed by a set of clauses $\Delta$; they are referred to as *1-empowering* with respect to $\Delta$. (The exact relationship between 1-empowering and absorption is discussed in Atserias et al., 2011.)

We will now explore the relationship between absorption and negative-hyper-resolution.

**Example 5** *Let $\Delta$ be the direct encoding of a CSP instance $P = (V, D, C)$, where $V = \{u, v, w\}$, $D_u = D_v = D_w = \{1, 2\}$ and $C$ contains two binary constraints: one forbids the assignment of the value 1 to $u$ and $v$ simultaneously, and the other forbids the simultaneous assignment of the value 2 to $u$ and 1 to $w$. Let $C$ also contain a ternary constraint that forbids the assignment of the value 2 to all three variables simultaneously.*

$$\Delta = \{ \ x_{u1} \vee x_{u2}, \ x_{v1} \vee x_{v2}, \ x_{w1} \vee x_{w2},$$
$$\neg x_{u1} \vee \neg x_{u2}, \ \neg x_{v1} \vee \neg x_{v2}, \ \neg x_{w1} \vee \neg x_{w2},$$
$$\neg x_{u1} \vee \neg x_{v1}, \ \neg x_{u2} \vee \neg x_{w1}, \ \neg x_{u2} \vee \neg x_{v2} \vee \neg x_{w2} \ \}.$$

*The clause $\neg x_{v1} \vee \neg x_{w1}$ is not contained in $\Delta$, but can be obtained by negative-hyper-resolution from the clauses $x_{u1} \vee x_{u2}, \neg x_{u1} \vee \neg x_{v1}, \neg x_{u2} \vee \neg x_{w1}$.*

*This clause is absorbed by $\Delta$, since every inconclusive round that sets $x_{v1} = true$ must set $x_{w1} = false$ by unit propagation, and every inconclusive round that sets $x_{w1} = true$ must set $x_{v1} = false$ also by unit propagation.*

Example 5 indicates that clauses that can be obtained by negative hyper-resolution from a set of clauses $\Delta$ are sometimes absorbed by $\Delta$. The next result clarifies when this situation holds.

**Lemma 4** *Any negative-hyper-resolvent of a set of disjoint clauses is absorbed by that set of clauses.*

*Proof.* Let $C$ be the negative-hyper-resolvent of a set of clauses $\Delta = \{C_i \vee \neg x_i \mid i = 1, 2, \ldots, r\}$ and a clause $C' = C_0 \vee x_1 \vee x_2 \vee \cdots \vee x_r$, where each $C_i$ is a (possibly empty) disjunction of negative literals, for $0 \leq i \leq r$. Then $C = C_0 \vee C_1 \vee \cdots \vee C_r$ by Definition 3. By Definition 4, we must show that $\Delta \cup C'$ absorbs $C$ at each of its literals. Assume all but one of the literals of $C$ are falsified. Since the set of clauses $\Delta \cup C'$ are assumed to be disjoint, the remaining literal $l$ must belong to exactly one of the clauses in this set. There are two cases to consider.

1. If $l$ belongs to the clause $C'$, then all clauses in $\Delta$ have all but one literals falsified, so the remaining literal $\neg x_i$ in each of these clauses is set to true, by unit propagation. Hence all literals in $C'$ are falsified, except for $l$, so $l$ is set to true, by unit propagation.

2. If $l$ belongs to one of the clauses $C_i \vee \neg x_i$, then all of the remaining clauses in $\Delta$ have all but one literals falsified, so the corresponding literals $\neg x_j$ are set to true, by unit propagation. Hence all literals in $C'$ are falsified, except for $x_i$, so $x_i$ is set to true, by unit propagation. But now all literals in $C_i \vee \neg x_i$ are falsified, except for $l$, so $l$ is set to true by unit propagation.

□

The next example shows that the negative-hyper-resolvent of a set of clauses that is *not* disjoint will *not* necessarily be absorbed by those clauses.

**Example 6** *Recall the set of clauses $\Delta$ given in Example 5, which is the direct encoding of a CSP instance with three variables $\{u, v, w\}$, each with domain $\{1, 2\}$.*

*The clause $\neg x_{u2} \vee \neg x_{v2}$ is not contained in $\Delta$, but can be obtained by negative-hyper-resolution from the clauses $x_{w1} \vee x_{w2}, \neg x_{u2} \vee \neg x_{v2} \vee \neg x_{w2}, \neg x_{u2} \vee \neg x_{w1}$.*

*This clause is not absorbed by $\Delta$, since an inconclusive round that sets $x_{v2} = true$ will not necessarily ensure that $x_{u2} = false$ by unit propagation.*

The basic approach we shall use to establish our main results below is to show that any clauses that can be obtained by bounded width negative-hyper-resolution from a given set of clauses, but are not immediately absorbed (such as the one in Example 6) are likely to become absorbed quite quickly because of the additional clauses that are added by the process of clause learning. Hence a clause-learning SAT-solver is likely to fairly rapidly absorb all of the clauses that can be derived from its original database of clauses by negative-hyper-resolution. In particular, if the empty clause can be derived by negative-hyper-resolution, then the solver will fairly rapidly absorb some literal and its complement, and hence report unsatisfiability (see the proof of Theorem 2 for details).

The following key properties of absorption are established by Atserias et al. (2011).

**Lemma 5** *(Atserias et al., 2011) Let $\Delta$ and $\Delta'$ be sets of clauses, and let $C$ and $C'$ be non-empty clauses.*

1. *If $C$ belongs to $\Delta$, then $\Delta$ absorbs $C$;*

2. *If $C \subseteq C'$ and $\Delta$ absorbs $C$, then $\Delta$ absorbs $C'$;*

3. *If $\Delta \subseteq \Delta'$ and $\Delta$ absorbs $C$, then $\Delta'$ absorbs $C$.*

To allow further analysis, we need to make some assumptions about the *learning scheme*, the *restart policy* and the *branching strategy* used by our SAT-solver.

The learning scheme is a rule that creates and adds a new clause to the database whenever there is a conflict. Such a clause is called a *conflict clause*, and each of its literals is falsified by some assignment in the current state. If a literal is falsified by the $i$-th decision assignment, or some later implied assignment before the $(i+1)$-th decision assignment, it is said to be *falsified* at level $i$. If a conflict clause contains exactly one literal that is falsified at the maximum possible level, it is called an *asserting clause* (Pipatsrisawat & Darwiche, 2009; Zhang, Madigan, Moskewicz, & Malik, 2001).

**Assumption 1** *The learning scheme chooses an* asserting clause.

---

**Algorithm 1** Framework for a typical clause-learning SAT-solver

---

**Input:** $\Delta$ : set of clauses;
  $S$ : partial assignment of truth values to variables.

1. **while** $\Delta|_S \neq \emptyset$ **do**
2.   **if** $\emptyset \in \Delta|_S$ **then**                                    CONFLICT
3.     **if** $S$ contains no decision assignments **then**
4.       **print** "UNSATISFIABLE" and halt
5.     **else**
6.       apply the *learning scheme* to add a new clause to $\Delta$
7.       **if** *restart policy* says restart **then**
8.         set $S = \emptyset$
9.       **else**
10.        select most recent conflict-causing unreversed decision assignment in $S$
11.        reverse this decision, and remove all later assignments from $S$
12.      **end if**
13.    **end if**
14.  **else if** $\{l\} \in \Delta|_S$ for some literal $l$ **then**               UNIT PROPAGATION
15.    add to $S$ the *implied assignment* $x = a$ which satisfies $l$
16.  **else**                                                        DECISION
17.    apply the *branching strategy* to choose a *decision assignment* $x = a$
18.    add this decision assignment to $S$
19.  **end if**
20. **end while**
21. **print** "SATISFIABLE" and output $S$

---

Most learning schemes in current use satisfy this assumption (Pipatsrisawat & Darwiche, 2009; Zhang et al., 2001), including the learning schemes called "1UIP" and "Decision" (Zhang et al., 2001).

We make no particular assumption about the restart policy. However, our main result is phrased in terms of a bound on the expected number of restarts. If the algorithm restarts after $r$ conflicts, our bound on the expected number of restarts can simply be multiplied by $r$ to get a bound on the expected number of conflicts. This means that the results will be strongest if the algorithm restarts *immediately after each conflict*. In that case, $r = 1$ and our bound will also bound the expected number of conflicts. Existing SAT-solvers typically do not employ such an aggressive restart policy, but we note the remark in the work of Pipatsrisawat and Darwiche (2009, p.666) that "there has been a clear trend towards more and more frequent restarts for modern SAT solvers".

The branching strategy determines which decision assignment is chosen after an inconclusive round that is not complete. In most current SAT solvers the strategy is based on some heuristic measure of *variable activity*, which is related to the occurrence of a variable in conflict clauses (Moskewicz, Madigan, Zhao, Zhang, & Malik, 2001). However, to simplify the probabilistic analysis, we will make the following assumption.

**Assumption 2** *The branching strategy chooses a variable uniformly at random amongst the unassigned variables, and assigns it the value TRUE.*

As noted by Atserias et al. (2011), the same analysis we give below can also be applied to any other branching strategy that randomly chooses between making a heuristic-based decision or a randomly-based decision. More precisely, if we allow, say, $c > 1$ rounds of non-random decisions between random ones, then the number of required restarts and conflicts would appear multiplied by a factor of $c$.

An algorithm that behaves according to the description in Algorithm 1, and satisfies the assumptions above, will be called a *standard randomised* SAT-solver.

**Theorem 2** *If a set of non-empty clauses $\Delta$ over $n$ Boolean variables has a negative-hyper-resolution refutation of width $k$ and length $m$, then the expected number of restarts required by a standard randomised SAT-solver to discover that $\Delta$ is unsatisfiable is less than $mnk^2\binom{n}{k}$.*

*Proof.* Let $C_1, C_2, \ldots, C_m$ be a negative-hyper-resolution refutation of width $k$ from $\Delta$, where $C_m$ is the first occurrence of the empty clause. Since each clause in $\Delta$ is non-empty, $C_m$ must be derived by negative-hyper-resolution from some collection of negative literals $\neg x_1, \neg x_2, \ldots \neg x_d$ and a purely positive clause $x_1 \vee x_2 \vee \cdots \vee x_d \in \Delta$.

Now consider a standard randomised SAT-solver started with database $\Delta$. Once all of the unit clauses $\neg x_i$ are absorbed by the current database, then, by Definition 4, any further inconclusive round of the algorithm must assign all variables $x_i$ false, and hence falsify the clause $x_1 \vee x_2 \vee \cdots x_d$. Since this happens even when no decision assignments are made, the SAT-solver will report unsatisfiability.

It only remains to bound the expected number of restarts required until each clause $C_i$ is absorbed, for $1 \leq i < m$. Let each $C_i$ be the negative-hyper-resolvent of clauses $C_{i1}, C_{i2}, \ldots, C_{ir}$, each of the form $C'_{ij} \vee \neg x_j$, together with a clause $C_{i0} = C_0 \vee x_1 \vee x_2 \vee \cdots \vee x_r$ from $\Delta$, where $C_0$ is a (possibly empty) disjunction of negative literals. Assume also that each clause $C_{ij}$ is absorbed by $\Delta$ for $j = 0, 1, \ldots, r$.

If $\Delta$ absorbs $C_i$, then no further learning or restarts are needed, so assume now that $\Delta$ does not absorb $C_i$. By Definition 4, this means that there exists some literal $l$ and some inconclusive round $R$ started with $\Delta$ that falsifies $C_i \setminus \{l\}$ and does not satisfy $C_i$. Note that $R$ must leave the literal $l$ unassigned, because one assignment would satisfy $C_i$ and the other would falsify $C_0$ and each $C'_{ij}$, and hence force all of the literals $\neg x_j$ used in the negative-hyper-resolution step to be satisfied, because each $C_{ij}$ is absorbed by $\Delta$, so $C_{i0}$ would be falsified, contradicting the fact that $R$ is inconclusive.

Hence, if the branching strategy chooses to falsify the literals $C_i \setminus \{l\}$ whenever it has a choice, it will construct an inconclusive round $R'$ where $l$ is unassigned (since all the decision assignments in $R'$ are also assigned the same values in $R$, any implied assignments in $R'$ must also be assigned the same values[3] in $R$, but we have shown that $R$ leaves $l$ unassigned). If the branching strategy then chooses to falsify the remaining literal $l$ of $C_i$, then the algorithm would construct a conclusive round $R''$ where $C_{i0}$ is falsified, and all decision assignments falsify literals in $C_i$. Hence, by Assumption 1, the algorithm would then learn some asserting clause $C'$ and add it to $\Delta$ to obtain a new set $\Delta'$.

---

3. See Lemmas 5, 8 and 10 in the work of Atserias et al. (2011) for a more formal statement and proof of this assertion.

Since $C'$ is an asserting clause, it contains exactly one literal, $l'$, that is falsified at the highest level in $R''$. Hence, any inconclusive round $R$ started with $\Delta'$ that falsifies $C_i \setminus \{l\}$ will falsify all but one literal of $C'$, and hence force the remaining literal $l'$ to be satisfied, by unit propagation. If this new implied assignment for $l'$ propagates to force $l$ to be true, then $R$ satisfies $C_i$, and hence $\Delta'$ absorbs $C_i$ at $l$. If not, then the branching strategy can once again choose to falsify the remaining literal $l$ of $C_i$, which will cause a new asserting clause to be learned and added to $\Delta$. Since each new asserting clause forces a new literal to be satisfied after falsifying $C_i \setminus \{l\}$ this process can be repeated fewer than $n$ times before it is certain that $\Delta'$ absorbs $C_i$ at $l$.

Now consider any sequence of $k$ random branching choices. If the first $k-1$ of these each falsify a literal of $C_i \setminus \{l\}$, and the final choice falsifies $l$, then we have shown that the associated round will reach a conflict, and add an asserting clause to $\Delta$. With a random branching strategy, as described in Assumption 2, the probability that this happens is at least the probability that the first $k-1$ random choices consist of a fixed set of variables (in some order), and the final choice is the variable associated with $l$. The number of random choices that fall in a fixed set follows the hypergeometric distribution, so the overall probability of this is $\frac{1}{\binom{n}{k-1}} \frac{1}{(n-k+1)} = 1/(k\binom{n}{k})$.

To obtain an upper bound on the expected number of restarts, consider the worst case where we require $n$ asserting clauses to be added to absorb each clause $C_i$ at each of its $k$ literals $l$. Since we require only an upper bound, we will treat each round as an independent trial with success probability $p = 1/(k\binom{n}{k})$, and consider the worst case where we have to achieve $(m-1)nk$ successes to ensure that $C_i$ for $1 \leq i < m$ is absorbed. In this case the total number of restarts will follow a negative binomial distribution, with expected value $(m-1)nk/p$. Hence in all cases the expected number of restarts is less than $mnk^2\binom{n}{k}$. □

A tighter bound on the number of restarts can be obtained if we focus on the DECISION learning scheme (Atserias et al., 2011; Zhang et al., 2001), as the next result indicates.

**Theorem 3** *If a set of non-empty clauses $\Delta$ over $n$ Boolean variables has a negative-hyper-resolution refutation of width $k$ and length $m$, then the expected number of restarts required by a standard randomised SAT-solver using the DECISION learning scheme to discover that $\Delta$ is unsatisfiable is less than $m\binom{n}{k}$.*

*Proof.* The proof is similar to the proof of Theorem 2, except that the DECISION learning scheme has the additional feature that the literals in the chosen conflict clause falsify a subset of the current decision assignments. Hence in the situation we consider, where the decision assignments all falsify literals of some clause $C_i$, this learning scheme will learn a subset of $C_i$, and hence immediately absorb $C_i$, by Lemma 5 (1,2). Hence the maximum number of learnt clauses required is reduced from $(m-1)nk$ to $(m-1)$, and the probability is increased from $1/(k\binom{n}{k})$ to $1/\binom{n}{k}$, giving the tighter bound. □

Note that a similar argument shows that the standard deviation of the number of restarts is less than the standard deviation of a negative binomial distribution with parameters $m$ and $1/\binom{n}{k}$, which is less than $\sqrt{m}\binom{n}{k}$. Hence, by Chebyshev's inequality (one-tailed version), the probability that a standard randomised SAT-solver using the DECISION learning scheme will discover that $\Delta$ is unsatisfiable after $(m + \sqrt{m})\binom{n}{k}$ restarts is greater than $1/2$.

## 5. $k$-Consistency and SAT-Solvers

By combining Theorem 1 and Theorem 3 we obtain the following result linking $k$-consistency and SAT-solvers.

**Theorem 4** *If the $k$-consistency closure of a CSP instance $P$ is empty, then the expected number of restarts required by a standard randomised SAT-solver using the DECISION learning scheme to discover that the direct encoding of $P$ is unsatisfiable is $O(n^{2k}d^{2k})$, where $n$ is the number of variables in $P$ and $d$ is the maximum domain size.*

*Proof.* The length $m$ of a negative-hyper-resolution refutation of width $k$ is bounded by the number of possible no-goods of length at most $k$ for $P$, which is $\sum_{i=1}^{k} d^i \binom{n}{i}$. Hence, by Theorem 1 and Theorem 3 we obtain a bound of $\left( \sum_{i=1}^{k} d^i \binom{n}{i} \right) \binom{nd}{k}$, which is $O(n^{2k}d^{2k})$. □

Hence a standard randomised SAT-solver with a suitable learning strategy will decide the satisfiability of any CSP instance with tree-width $k$ with $O(n^{2k}d^{2k})$ expected restarts, even when it is set to restart immediately after each conflict. In particular, the satisfiability of any tree-structured binary CSP instance (i.e., with tree-width 1) will be decided by such a solver with at most $O(n^2d^2)$ expected conflicts, which is comparable with the growth rate of an optimal arc-consistency algorithm for binary constraints. Note that this result cannot be obtained directly from the work of Atserias et al. (2011), because the direct encoding of an instance with tree-width $k$ is a set of clauses whose tree-width may be as high as $dk$.

Moreover, a standard randomised SAT-solver will decide the satisfiability of any CSP instance, with any structure, within the same polynomial bounds, if the constraint relations satisfy certain algebraic properties that ensure bounded width (Barto & Kozik, 2009). Examples of such constraint types include the "0/1/all" relations, defined by Cooper et al. (1994), and the "connected row-convex" relations, defined by Deville et al. (1997), which can both be decided by 2-consistency.

It was shown by Gent (2002) that the support encoding of a binary CSP instance can be made arc-consistent (that is, 1-consistent) by applying unit propagation alone. Hence, a standard SAT-solver will mimic the effect of enforcing arc-consistency on such an encoding before making any decisions or restarts. By combining Theorem 4 with the observation in Example 4 that the direct encoding can be obtained from the support encoding by negative-hyper-resolution, we obtain the following corollary concerning the support encoding for all higher levels of consistency.

**Corollary 2** *For any $k \geq 2$, if the $k$-consistency closure of a binary CSP instance $P$ is empty, then the expected number of restarts required by a standard randomised SAT-solver using the DECISION learning scheme to discover that the support encoding of $P$ is unsatisfiable is $O(n^{2k}d^{2k})$, where $n$ is the number of variables in $P$ and $d$ is the maximum domain size.*

The CSP literature describes many variations on the notion of consistency. In this paper we have considered $k$-consistency only. We note that our results can be generalised to some other types of consistency such as singleton arc-consistency (Bessière, 2006). The extension to singleton arc-consistency follows from the recent discovery that if a family of CSP instances is solvable by enforcing singleton arc-consistency, then the instances have bounded width (Chen, Dalmau, & Grußien, 2011). In other words, all such instances can be solved by enforcing $k$-consistency, for some fixed $k$. Hence, by Theorem 4, they will be solved in polynomial expected time by a standard randomised SAT-solver.

## 6. Experimental Results

The polynomial upper bounds we obtain in this paper are not asymptotic, they apply for all values of $n, m$ and $k$. However, they are very conservative, and are likely to be met very easily in practice.

To investigate how an existing SAT-solver actually performs, we measured the runtime of the MINISAT solver (Eén & Sörensson, 2003), version 2.2.0, on a family of CSP instances that can be decided by a fixed level of consistency. For comparison, we also ran our experiments on two state-of-the-art constraint solvers: we used MINION (Gent, Jefferson, & Miguel, 2006), version 0.12, and the G12 finite domain solver (Nethercote et al., 2007), version 1.4.

To match the simplified assumptions of our analysis more closely, we ran a further set of experiments on a core version of MINISAT in order to get a solver that uses only unit propagation and conflict-directed learning with restarts. We also modified the solver to follow the random branching strategy described above. Our solver does not delete any learnt clauses and uses an extreme restart policy that makes it restart whenever it encounters a conflict. It uses the same learning scheme as MINISAT. We refer to this modified solver as SIMPLE-MINISAT.

As the characteristic feature of the instances tested is their relatively low tree-width, we also used the TOULBAR2 solver (Sanchez et al., 2008). This solver implements the BTD (Backtracking with Tree-Decomposition) technique which has been shown to be efficient in practice, in contrast to earlier methods that had been proposed to attempt to exploit tree-decompositions of the input problem (Jégou & Terrioux, 2003). As the problem of finding a tree-decomposition of minimal width (i.e., the tree-width) is NP-hard, the BTD technique uses some approximations (described in Jégou & Terrioux, 2003). We note here that TOULBAR2 is designed for solving optimization problems, namely weighted CSPs, or WCSPs. In a WCSP instance, certain partial assignments have an associated cost. However, the TOULBAR2 solver can be used to solve standard CSPs by simply setting all costs to 0.

For all of the results, the times given are elapsed times on a Lenovo 3000 N200 laptop with an Intel Core 2 Duo processor running at 1.66GHz with 2GB of RAM. Each generated instance was run five times and the mean times and mean number of restarts are shown[4].

**Example 7** *We consider a family of instances specified by two parameters, $w$ and $d$. They have $((d-1)*w+2)*w$ variables arranged in groups of size $w$, each with domain $\{0, ..., d-1\}$.*

---

4. MINISAT and SIMPLE-MINISAT were run with different seeds for each of the five runs of an instance. Instances marked with * were run once only. The runtime of SIMPLE-MINISAT on those instances exceeded 6 hours. Moreover, TOULBAR2 was run with parameter $B = 1$ which enables BTD.

We impose a constraint of arity $2w$ on each pair of successive groups, requiring that the sum of the values assigned to the first of these two groups should be strictly smaller than the sum of the values assigned to the second. This ensures that the instances generated are unsatisfiable. An instance with $w = 2$ and $d = 2$ is shown diagrammatically and defined using the specification language MiniZinc (Nethercote et al., 2007) in Figure 1 (a) and (b) respectively[5]. A similar format is used for TOULBAR2 [6] and the same instance encoded in this format is shown in Figure 1 (c) (note that each hard constraint has cost 0).

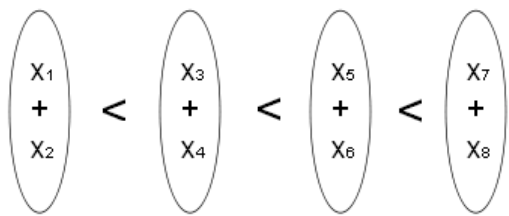

(a) Graphical representation.

array[1..4] of var 0..1 : $X1$;
array[1..4] of var 0..1 : $X2$;
constraint
forall($i$ in 1..3)(
$X1[i] + X2[i] < X1[i+1] + X2[i+1]$);
solve satisfy;

(b) Specification in MiniZinc.

chain
$x1$ 0 1
$x2$ 0 1
$x3$ 0 1
$x4$ 0 1
$x5$ 0 1
$x6$ 0 1
$x7$ 0 1
$x8$ 0 1
hard( $x1 + x2 < x3 + x4$ )
hard( $x3 + x4 < x5 + x6$ )
hard( $x5 + x6 < x7 + x8$ )

(c) Specification in $cp$ format.

Figure 1: An example of a CSP instance with $w = 2$, $d = 2$ and tree-width = 3.

The structure of the instances described in Example 7 has a simple tree-decomposition as a path of nodes, with each node corresponding to a constraint scope. Hence the tree-width of these instances is $2w - 1$ and they can be shown to be unsatisfiable by enforcing $(2w - 1)$-consistency (Atserias et al., 2007). However, these instances cannot be solved efficiently using standard propagation algorithms which only prune individual domain values.

The structure of the direct encoding of these instances also has a tree-decomposition with each node corresponding to a constraint scope in the original CSP instance. However, because the direct encoding introduces $d$ Boolean variables to represent each variable in the original instance, the tree-width of the encoded SAT instances is larger by approximately a factor of $d$; it is in fact $2wd - 1$ (see Figure 2).

---

5. In order to run an instance on a CP solver one must usually use a translator to convert the original model. The MiniZinc distribution provides an MZN2FZN translator while for MINION one can use TAILOR (available at http://www.cs.st-andrews.ac.uk/~andrea/tailor/).

6. A CP2WCSP translator and a description of the $cp$ and $wcsp$ formats is available at http://carlit.toulouse.inra.fr/cgi-bin/awki.cgi/SoftCSP.

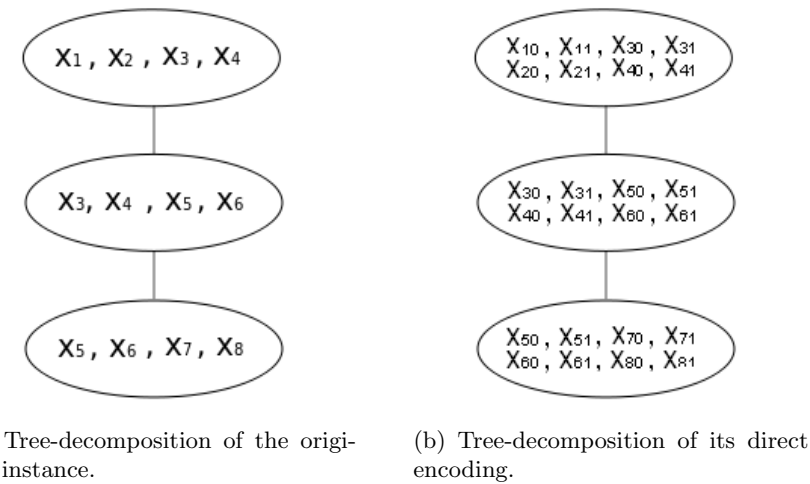

(a) Tree-decomposition of the original instance.

(b) Tree-decomposition of its direct encoding.

Figure 2: Tree-decompositions of the CSP instance from Figure 1.

Table 1 shows the runtimes of SIMPLE-MINISAT and the original MINISAT solver on this family of instances, along with times for the two state-of-the-art CP solvers and the WCSP solver TOULBAR2. By far the best solver for this set of instances is TOULBAR2, which is explicitly designed to exploit low tree-width by constructing a tree-decomposition. For the class of instances we are considering, the widths of the tree-decompositions found by TOULBAR2 matched the tree-widths of the instances tested (i.e., $2w - 1$).

However, we also note that MINISAT is remarkably effective in solving these chains of inequalities, compared to MINION and G12, even though the use of MINISAT requires encoding each instance into a large number of clauses with a much larger tree-width than the original. Although our simplified version of the MINISAT solver takes a little longer than the current highly optimised version, it still performs very well on these instances in comparison with the conventional CP solvers. Moreover, the number of restarts (and hence the number of conflicts) appears to grow only polynomially with the size of the instance (see Figure 3). In all cases the actual number of restarts is much lower than the polynomial upper bound on the expected number of restarts given in Theorem 4.

Our best theoretical upper bounds on the expected run-time were obtained for the DECISION learning scheme (Theorem 4), but the standard version of MINISAT uses the 1UIP learning scheme with *conflict clause minimization*. To allow a direct comparison with these theoretical upper bounds, we implemented the DECISION scheme in SIMPLE-MINISAT. As the 1UIP learning scheme has generally been found to be more efficient in practice (Zhang et al., 2001), we switched off conflict clause minimization in SIMPLE-MINISAT in order to compare the two standard learning schemes and ran a further set of experiments. We counted the number of restarts for these two modified solvers on instances of the form described in Example 7 - see Table 2.

| group size (w) | domain size (d) | CSP variables (n) | Minion (sec) | G12 (sec) | Toulbar2 (sec) | MiniSAT (sec) | simple-MiniSAT (sec) | simple-MiniSAT restarts |
|---|---|---|---|---|---|---|---|---|
| 2 | 2 | 8 | 0.055 | 0.010 | 0.021 | 0.003 | 0.002 | 19 |
| 2 | 3 | 12 | 0.053 | 0.011 | 0.023 | 0.005 | 0.007 | 157 |
| 2 | 4 | 16 | 0.057 | 0.013 | 0.040 | 0.015 | 0.034 | 820 |
| 2 | 5 | 20 | 0.084 | 0.047 | 0.091 | 0.043 | 0.188 | 3 039 |
| 2 | 6 | 24 | 1.048 | 0.959 | 0.199 | 0.126 | 0.789 | 7 797 |
| 2 | 7 | 28 | 47.295 | 122.468 | 0.549 | 0.362 | 2.884 | 17 599 |
| 2 | 8 | 32 | > 20 min | > 20 min | 1.214 | 0.895 | 9.878 | 36 108 |
| 2 | 9 | 36 | > 20 min | > 20 min | 2.523 | 2.407 | 34.352 | 65 318 |
| 2 | 10 | 40 | > 20 min | > 20 min | 4.930 | 5.656 | 111.912 | 114 827 |
| 3 | 2 | 15 | 0.055 | 0.010 | 0.024 | 0.004 | 0.008 | 167 |
| 3 | 3 | 24 | 0.412 | 0.034 | 0.103 | 0.066 | 0.503 | 5 039 |
| 3 | 4 | 33 | > 20 min | 7.147 | 0.860 | 1.334 | 20.054 | 41 478 |
| 3 | 5 | 42 | > 20 min | > 20 min | 5.646 | 20.984 | 817.779 | 210 298 |
| 3 | 6 | 51 | > 20 min | > 20 min | 28.663 | 383.564 | > 20 min | 731 860 |
| 4 | 2 | 24 | 0.060 | 0.015 | 0.046 | 0.012 | 0.118 | 1 617 |
| 4 | 3 | 40 | > 20 min | 11.523 | 1.246 | 4.631 | 260.656 | 108 113 |
| 4 | 4 | 56 | > 20 min | > 20 min | 20.700 | 1,160.873 | > 20 min | 1 322 784* |

Table 1: Average performance of solvers on instances from Example 7.

| group size (w) | domain size (d) | CSP variables (n) | no. of clauses in the direct encoding | simple-MiniSAT 1UIP (sec) | simple-MiniSAT 1UIP restarts | simple-MiniSAT Decision (sec) | simple-MiniSAT Decision restarts |
|---|---|---|---|---|---|---|---|
| 2 | 2 | 8 | 49 | 0.002 | 21 | 0.002 | 23 |
| 2 | 3 | 12 | 298 | 0.008 | 203 | 0.010 | 267 |
| 2 | 4 | 16 | 1 162 | 0.048 | 1 026 | 0.057 | 1 424 |
| 2 | 5 | 20 | 3 415 | 0.272 | 4 068 | 0.323 | 5 283 |
| 2 | 6 | 24 | 8 315 | 1.399 | 12 029 | 1.526 | 14 104 |
| 2 | 7 | 28 | 17 724 | 5.780 | 27 356 | 6.035 | 33 621 |
| 2 | 8 | 32 | 34 228 | 24.417 | 56 193 | 20.436 | 64 262 |
| 2 | 9 | 36 | 61 257 | 95.278 | 109 862 | 69.144 | 113 460 |
| 2 | 10 | 40 | 103 205 | 309.980 | 199 399 | 207.342 | 190 063 |
| 3 | 2 | 15 | 198 | 0.009 | 192 | 0.012 | 287 |
| 3 | 3 | 24 | 3 141 | 0.643 | 5 952 | 0.750 | 7 308 |
| 3 | 4 | 33 | 23 611 | 53.067 | 63 952 | 71.778 | 91 283 |
| 3 | 5 | 42 | 113 406 | 2,266.627 | 375 849 | 2,036.456 | 391,664 |
| 3 | 6 | 51 | 408 720 | > 6 hours | 1 584 012* | > 6 hours | 1 365 481* |
| 4 | 2 | 24 | 863 | 0.141 | 1 937 | 0.192 | 2 592 |
| 4 | 3 | 40 | 34 666 | 603.241 | 155 842 | 938.836 | 253 153 |

Table 2: Average performance of simple-MiniSAT with the 1UIP and the Decision learning schemes on instances from Example 7.

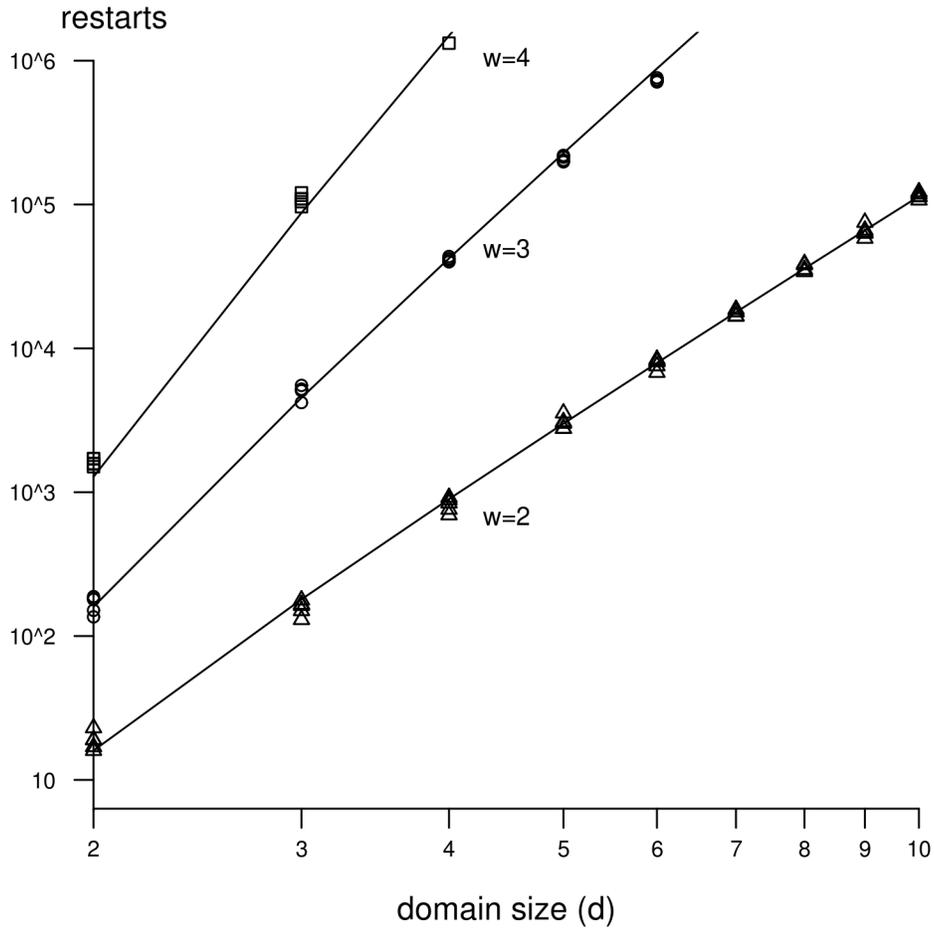

Figure 3: Log-log plot of the number of restarts/conflicts used by SIMPLE-MINISAT on the instances from Example 7. The solid lines show a growth function of $d^{2w-2}\binom{nd/w}{3}$, where $n$ is the number of CSP variables. This empirically derived polynomial function appears to fit the experimental data well, and is much lower than the upper bound on the expected number of restarts calculated in Theorem 4 which is $O(d^{4w-2}n^{4w-2})$.

Although the performance of SIMPLE-MINISAT with the DECISION learning scheme and the 1UIP scheme are significantly worse than the performance of the original SIMPLE-MINISAT solver, only about twice as many restarts were required for each instance. Hence, our theoretical upper bounds are still easily met for both of these standard learning schemes.

## 7. Conclusions

We have shown that the notion of $k$-consistency can be precisely captured by a single inference rule on the direct encoding of a CSP instance, restricted to deriving only clauses with at most $k$ literals. We used this to show that a clause-learning SAT-solver with a purely random branching strategy will simulate the effect of enforcing $k$-consistency in expected polynomial time, for all fixed $k$. This is sufficient to ensure that such solvers are able to solve certain problem families much more efficiently than conventional CP solvers relying on GAC-propagation.

In principle clause-learning SAT-solvers can also do much more. It is known that, with an appropriate branching strategy and restart policy, they are able to p-simulate general resolution (Beame et al., 2004; Pipatsrisawat & Darwiche, 2009), and general resolution proofs can be exponentially shorter than the negative-hyper-resolution proofs we have considered here (Hwang & Mitchell, 2005). In practice, it seems that current clause-learning SAT-solvers with highly-tuned learning schemes, branching strategies and restart policies are often able to exploit structure in the Boolean encoding of a CSP instance even more effectively than local consistency techniques. Hence considerable work remains to be done in understanding the relevant features of instances which they are able to exploit, in order to predict their effectiveness in solving different kinds of CSP instances.

## Acknowledgments

We would like to thank Albert Atserias and Marc Thurley for comments on the conference version of this paper, as well as the anonymous referees. The provision of an EPSRC Doctoral Training Award to Justyna Petke is also gratefully acknowledged.

A preliminary version of this paper appeared in *Proceedings of the $16^{th}$ International Conference on Principles and Practice of Constraint Programming - CP2010*.

## References

Atserias, A., Bulatov, A. A., & Dalmau, V. (2007). On the power of $k$-consistency. In *International Colloquium on Automata, Languages and Programming - ICALP'07*, pp. 279–290.

Atserias, A., & Dalmau, V. (2008). A combinatorial characterization of resolution width. *Journal of Computer and System Sciences*, *74*(3), 323–334.

Atserias, A., Fichte, J. K., & Thurley, M. (2011). Clause-learning algorithms with many restarts and bounded-width resolution. *Journal of Artificial Intelligence Research (JAIR)*, *40*, 353–373.

Bacchus, F. (2007). GAC via unit propagation. In *Principles and Practice of Constraint Programming - CP'07*, pp. 133–147.

Barto, L., & Kozik, M. (2009). Constraint satisfaction problems of bounded width. In *Symposium on Foundations of Computer Science - FOCS'09*, pp. 595–603.

Beame, P., Kautz, H. A., & Sabharwal, A. (2004). Towards understanding and harnessing the potential of clause learning. *Journal of Artificial Intelligence Research (JAIR)*, *22*, 319–351.

Bessière, C. (2006). Constraint propagation. In Rossi, F., van Beek, P., & Walsh, T. (Eds.), *Handbook of Constraint Programming*, chap. 3. Elsevier.

Büning, H., & Lettmann, T. (1999). *Propositional logic: deduction and algorithms.* Cambridge tracts in theoretical computer science. Cambridge University Press.

Chen, H., Dalmau, V., & Grußien, B. (2011). Arc consistency and friends. *Computing Research Repository - CoRR, abs/1104.4993*.

Cooper, M. C. (1989). An optimal $k$-consistency algorithm. *Artificial Intelligence*, *41*(1), 89–95.

Cooper, M. C., Cohen, D. A., & Jeavons, P. (1994). Characterising tractable constraints. *Artificial Intelligence*, *65*(2), 347–361.

de Kleer, J. (1989). A comparison of ATMS and CSP techniques. In *International Joint Conference on Artificial Intelligence - IJCAI'89*, pp. 290–296.

Deville, Y., Barette, O., & Hentenryck, P. V. (1997). Constraint satisfaction over connected row convex constraints. In *International Joint Conference on Artificial Intelligence - IJCAI'97 (1)*, pp. 405–411.

Eén, N., & Sörensson, N. (2003). An extensible SAT-solver. In *Theory and Applications of Satisfiability Testing - SAT'03*, pp. 502–518.

Freuder, E. C. (1978). Synthesizing constraint expressions. *Communications of the ACM*, *21*(11), 958–966.

Gent, I. P. (2002). Arc consistency in SAT. In *European Conference on Artificial Intelligence - ECAI'02*, pp. 121–125.

Gent, I. P., Jefferson, C., & Miguel, I. (2006). MINION: A fast scalable constraint solver. In *European Conference on Artificial Intelligence - ECAI'06*, pp. 98–102.

Hooker, J. N. (2006). *Integrated Methods for Optimization (International Series in Operations Research & Management Science).* Springer-Verlag New York, Inc., Secaucus, NJ, USA.

Hoos, H. H. (1999). SAT-encodings, search space structure, and local search performance. In *International Joint Conference on Artificial Intelligence - IJCAI'99*, pp. 296–303.

Hwang, J., & Mitchell, D. G. (2005). 2-way vs. d-way branching for CSP. In *Principles and Practice of Constraint Programming - CP'05*, pp. 343–357.

Jégou, P., & Terrioux, C. (2003). Hybrid backtracking bounded by tree-decomposition of constraint networks. *Artificial Intelligence*, *146*(1), 43–75.

Kolaitis, P. G., & Vardi, M. Y. (2000). A game-theoretic approach to constraint satisfaction. In *Conference on Artificial Intelligence - AAAI'00 / Innovative Applications of Artificial Intelligence Conference - IAAI'00*, pp. 175–181.

Mackworth, A. K. (1977). Consistency in networks of relations. *Artificial Intelligence*, *8*(1), 99–118.

Montanari, U. (1974). Networks of constraints: Fundamental properties and applications to picture processing. *Information Sciences*, *7*, 95–132.

Moskewicz, M. W., Madigan, C. F., Zhao, Y., Zhang, L., & Malik, S. (2001). Chaff: Engineering an efficient SAT solver. In *Design Automation Conference - DAC'01*, pp. 530–535.

Nethercote, N., Stuckey, P. J., Becket, R., Brand, S., Duck, G. J., & Tack, G. (2007). MiniZinc: Towards a standard CP modelling language. In *Principles and Practice of Constraint Programming - CP'07*, pp. 529–543.

Petke, J., & Jeavons, P. (2009). Tractable benchmarks for constraint programming. *Technical Report RR-09-07, Department of Computer Science, University of Oxford*.

Pipatsrisawat, K., & Darwiche, A. (2009). On the power of clause-learning SAT solvers with restarts. In *Principles and Practice of Constraint Programming - CP'09*, pp. 654–668.

Prestwich, S. D. (2009). CNF encodings. In Biere, A., Heule, M., van Maaren, H., & Walsh, T. (Eds.), *Handbook of Satisfiability*, pp. 75–97. IOS Press.

Rish, I., & Dechter, R. (2000). Resolution versus search: Two strategies for SAT. *Journal of Automated Reasoning*, *24*(1/2), 225–275.

Robinson, J. A. (1965). A machine-oriented logic based on the resolution principle. *Journal of the ACM*, *12*(1), 23–41.

Sanchez, M., Bouveret, S., de Givry, S., Heras, F., Jégou, P., Larrosa, J., Ndiaye, S., Rollon, E., Schiex, T., Terrioux, C., Verfaillie, G., & Zytnicki, M. (2008). Max-CSP competition 2008: Toolbar2 solver description. In *Proceedings of the Third International CSP Solver Competition*.

Schiex, T., & Verfaillie, G. (1993). Nogood recording for static and dynamic constraint satisfaction problems. In *International Conference on Tools with Artificial Intelligence - ICTAI'93*, pp. 48–55.

Tamura, N., Taga, A., Kitagawa, S., & Banbara, M. (2009). Compiling finite linear CSP into SAT. *Constraints*, *14*(2), 254–272.

van Dongen, M., Lecoutre, C., & Roussel, O. (2008). 3rd international CSP solver competition. Instances and results available at `http://www.cril.univ-artois.fr/CPAI08/`.

van Dongen, M., Lecoutre, C., & Roussel, O. (2009). 4th international CSP solver competition. Instances and results available at `http://www.cril.univ-artois.fr/CPAI09/`.

Walsh, T. (2000). SAT v CSP. In *Principles and Practice of Constraint Programming - CP'00*, pp. 441–456.

Zhang, L., Madigan, C. F., Moskewicz, M. W., & Malik, S. (2001). Efficient conflict driven learning in Boolean satisfiability solver. In *International Conference on Computer-Aided Design - ICCAD'01*, pp. 279–285.

Zhang, L., & Malik, S. (2002). The quest for efficient Boolean satisfiability solvers. In *Computer Aided Verification - CAV'02*, pp. 17–36.